\documentclass{article}

\usepackage[preprint]{neurips_2023}

\usepackage[utf8]{inputenc} 
\usepackage[T1]{fontenc}    
\usepackage{hyperref}       
\usepackage{url}            
\usepackage{booktabs}       
\usepackage{amsfonts}       
\usepackage{nicefrac}       
\usepackage{microtype}      
\usepackage{xcolor}         
\usepackage{graphicx}
\usepackage{makecell}
\usepackage{enumitem}
\usepackage{amssymb}
\usepackage{graphicx}
\usepackage{subfigure}
\usepackage[most]{tcolorbox}
\definecolor{softblue}{rgb}{0.8,0.9,1} 

\title{Multimodal ChatGPT for Medical Applications: an Experimental Study of GPT-4V}

\author{
    Zhiling Yan$^1$\footnotemark[1] \quad
    Kai Zhang$^1$\footnotemark[1] \quad
    Rong Zhou$^1$ \quad
    Lifang He$^1$ \quad
    Xiang Li$^2$ \quad
    Lichao Sun$^1 \footnotemark[2]$
    \\
    \\ $^{1}$Lehigh University, $^{2}$Massachusetts General Hospital and Harvard Medical School
}

\begin{document}

\maketitle

\begin{abstract}
In this paper, we critically evaluate the capabilities of the state-of-the-art multimodal large language model, i.e., GPT-4 with Vision (GPT-4V), on Visual Question Answering (VQA) task. Our experiments thoroughly assess GPT-4V's proficiency in answering questions paired with images using both pathology and radiology datasets from 11 modalities (e.g. Microscopy, Dermoscopy, X-ray, CT, etc.) and fifteen objects of interests (brain, liver, lung, etc.). Our datasets encompass a comprehensive range of medical inquiries, including sixteen distinct question types. Throughout our evaluations, we devised textual prompts for GPT-4V, directing it to synergize visual and textual information.
The experiments with accuracy score conclude that the current version of GPT-4V
is not recommended for real-world diagnostics due to its unreliable and suboptimal accuracy in
responding to diagnostic medical questions. In addition,
we delineate seven unique facets of GPT-4V's behavior in medical VQA, highlighting its constraints within this complex arena. The complete details of our evaluation cases are accessible at  \href{https://github.com/ZhilingYan/GPT4V-Medical-Report}{Github}.
\end{abstract}

\renewcommand{\thefootnote}{\fnsymbol{footnote}} 
\footnotetext[1]{These authors contributed equally to this work.} 
\footnotetext[2]{Corresponding author.} 

\renewcommand{\thefootnote}{}

\footnotetext{Lichao Sun (lis221@lehigh.edu)}

\renewcommand{\thefootnote}{\arabic{footnote}}

\begin{figure}[h]
  \centering
  \includegraphics[width=5.6in]{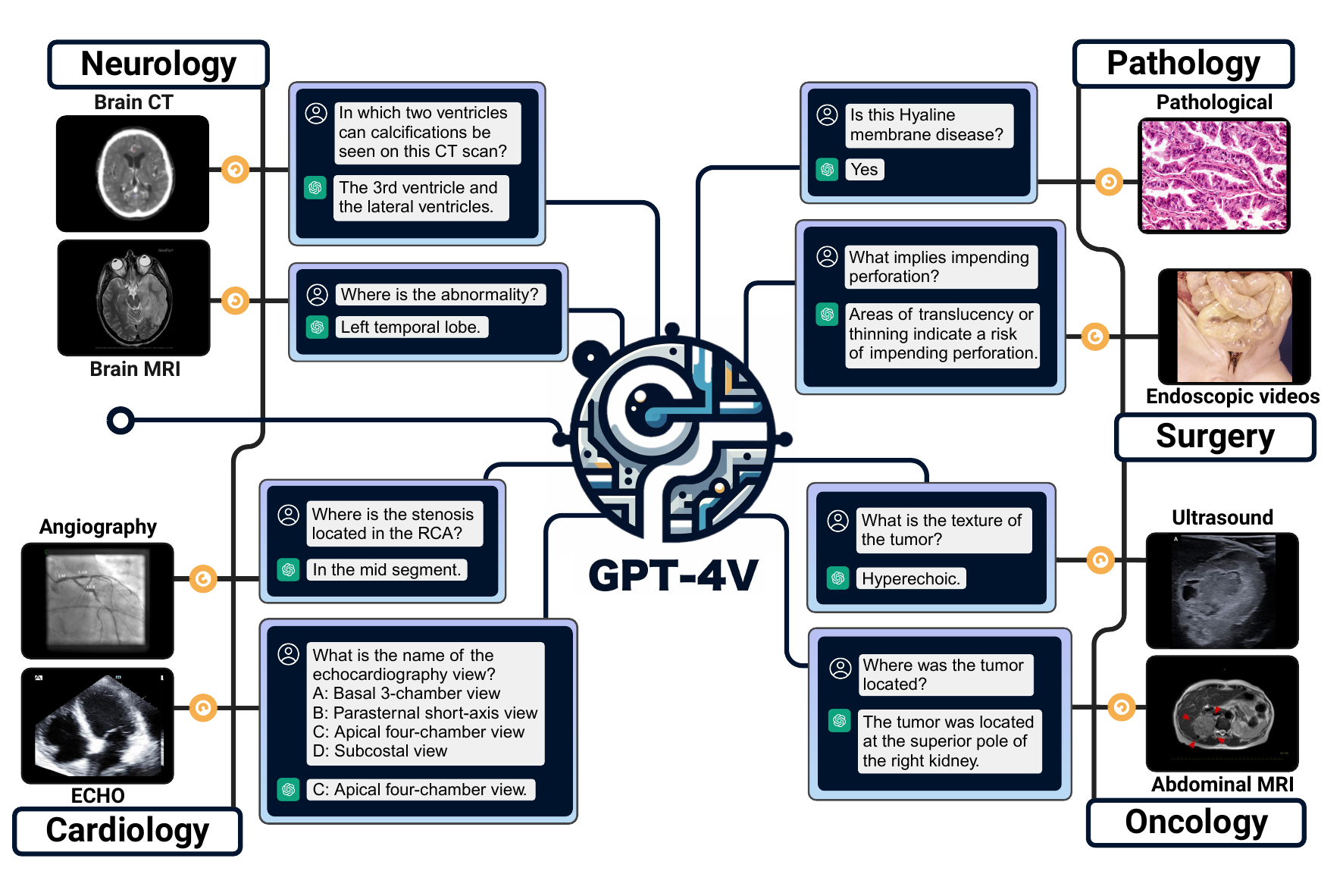}
  \caption{The diagram of medical departments and their corresponding objects of interest and modalities. We comprehensively consider 11 modalities across 15 objects of interest in the paper.}
  \label{fig: diagram of departments}
\end{figure}



\section{Introduction} 
\label{sec: intro}

Medicine is an intrinsically multimodal discipline. Clinicians often use medical images, clinical notes, lab tests, electronic health records, genomics, and more when providing care \citep{tu2023towards}. However, although many AI models have demonstrated great potential for biomedical applications~\citep{lee2023ai, nori2023capabilities, lee2023benefits}, majority of them focus on just one type of data \citep{acosta2022multimodal, singhal2023large, zhou2023attentive}. Some need extra adjustments to work well for specific tasks \citep{zhang2023biomedgpt, chen2023towards}, while others are not available for public use \citep{tu2023towards, li2023llava}. Previously, ChatGPT~\citep{openai2022gpt4} has shown potential in offering insights for both patients and doctors, aiding in minimizing missed or misdiagnoses~\citep{yunxiang2023chatdoctor, han2023medalpaca, wang2023huatuo}. A notable instance was when ChatGPT accurately diagnosed a young boy after 17 doctors had been unable to provide conclusive answers over three years\footnote{\url{https://www.today.com/health/mom-chatgpt-diagnosis-pain-rcna101843}}. The advent of visual input capabilities in the GPT-4 system, known as GPT-4 with Vision (GPT-4V)~\citep{openai2023gpt4v}, further amplifies this potential. This enhancement allows for more holistic information processing in a question and answer format, possibly mitigating language ambiguities—particularly relevant when non-medical experts, such as patients, might struggle with accurate descriptions due to unfamiliarity with medical jargon.

\subsection{Analysis Dimensions}

In this paper, we will systematically examine how GPT-4V operates in the medical field using the VQA approach. We believe this might become a predominant method for future medical AI like daily healthcare assistant. We will address questions in a hierarchical manner, emphasizing the step-by-step process of medical progression for an intelligent machine / agent. This approach will help us uncover what GPT-4V can offer and also delve into its limitations and challenges in real-world medical applications.

\begin{itemize}[left=0pt]
\item{In medical imaging, the fundamental aspect of machine understanding lies in recognizing the used modalities, such as X-Ray, MRI, or microscopy. Following this recognition, an effective system should discern specific objects within these images, from anatomical structures to distinct cellular configurations. The adeptness of GPT-4V in these preliminary tasks could offer insights into its proficiency in interpreting medical images across varied domains, setting the foundation for more advanced evaluations.}

\item{Central to medical analysis is the notion of localization, or the precise pinpointing of specific regions or objects within an image. Such capability is instrumental when demarcating between healthy and pathological tissues or discerning the positions of tumors, vessels, and other salient structures. For a model like GPT-4V, mastering this skill could usher in nuanced analyses, bolstering clinical diagnostics, treatment design, and disease tracking.}

\item{Further deepening the analysis, the precision with which GPT-4V gauges the dimensions of regions of interest (ROIs) becomes paramount. Monitoring the dynamics of tumors, evaluating organ dimensions, or quantifying lesions holds clinical weight, aiding in diagnostics, surgical planning, and gauging ailment severity.}

\item{Another layer of analytical depth involves the identification of morphological patterns. Such patterns—be it the systematic cellular structures in pathology or attributes such as density, form, and opacity in radiology—are instrumental for diagnostic deliberations. A case in point is the palisade-like cellular organization around necrotic zones, characteristic of glioblastoma multiforme, a specific brain malignancy.}

\item{Expanding the purview beyond mere visual cues, an integrative diagnostic modality combines imagery with textual descriptions, offering a holistic view of a patient's status. However, the efficacy of GPT-4V in such vision-language synthesis warrants exploration, particularly given concerns of potential over-reliance on singular modes, leading to possible incomplete or skewed diagnostic outcomes.}

\item{A mere answer, devoid of context or clarity, often falls short in the medical domain. Therefore, assessing if GPT-4V elucidates its rationale, articulates clearly, and evinces assurance in its responses becomes pivotal. Such a facet not only engenders trust among users but also aligns with the gravity and precision the medical domain demands.}

\item{Lastly, shaping the user-AI interaction framework remains crucial. Crafting an optimal prompt template when querying GPT-4V or similar AI entities can drastically influence response accuracy. Preliminary observations suggest that while GPT-4V's immediate answers might sometimes falter, certain prompt structures channel its analytical depth more effectively. Sharing such findings can guide users in their engagement with GPT-4V, optimizing outcomes in personal healthcare inquiries and consultations.}

\end{itemize}

\subsection{Highlights}
In this section, we provide a concise summary of our findings related to the characteristics of GPT-4V in the context of medical VQA. These characteristics, depicted in Section \ref{sec: features}, directly correspond to the research questions posed earlier:


\textbf{(1)} GPT-4V consistently recognizes various medical imaging modalities and the objects within them.

\textbf{(2)} For accurate localization, GPT-4V requires cues, particularly to consider the orientations of medical images across different modalities.

\textbf{(3)} GPT-4V finds it challenging to discern the size of Regions of Interest (ROI) or objects, especially when the assessment involves multiple slices, such as CT scans.


\textbf{(4)} While GPT-4V has the capability to integrate both image and text inputs for diagnostic-related queries, it displays tendencies towards visual and linguistic biases. Specifically, it might either overemphasize markings in images or rely excessively on text, neglecting the visual information in the process.

\textbf{(5)} GPT-4V typically offers cautious responses, emphasizing that it is not a medical professional (e.g., radiologist or pathologist). Nonetheless, its answers are thorough and come with detailed explanations. It's important to note that these explanations, while informative, are not definitive facts and should be cross-checked by experts for accuracy.

\textbf{(6)} Based on the statistical results concerning the accuracy of VQA, the current version of GPT-4V is not recommended for real-world diagnostics due to its unreliable and suboptimal accuracy in responding to diagnostic medical questions (see Section \ref{sec: results} for details).



\subsection{Contributions}
This report provides the following contributions to the community in the realm of medical AI:

\begin{itemize}[left=0pt,label=$\bigstar$]
    \item We meticulously assess GPT-4V's performance in responding to visually paired medical queries, leveraging datasets from a diverse array of seven imaging modalities such as Microscopy, Dermoscopy, X-ray, and CT, and centering our analysis on fifteen different clinical objects including the brain, liver, and lung. Our comprehensive dataset, uniquely tailored, encompasses sixteen different types of medical questions, providing a broad basis for evaluation.

\item The empirical results, derived from rigorous testing for accuracy, unequivocally suggest that the current version of GPT-4V should not be employed for practical diagnostic purposes. Its performance in responding to diagnostic medical questions demonstrates a lack of reliability and accuracy necessary for real-world application.

\item  Our study delineates seven distinct dimensions of GPT-4V's operational capabilities within the medical VQA context. These dimensions highlight the model's operational boundaries and shed light on its adaptability and limitations in the demanding realm of medical inquiry.
\end{itemize}


\section{Experimental Setup}
We outline the experimental setup or case studies employed to address the aforementioned questions and objectives. Since GPT-4V has not officially released APIs, we evaluated its capability for medical VQA through its dedicated chat interface (the ChatGPT webpage version), initiating our dialogue with image inputs. To eliminate any interference, hints, or biases from a multi-round conversation, we began a new chat session for each new Q\&A case. This ensured that GPT-4V did not unintentionally reference information from previous conversations related to different cases. In this report, we prioritize evaluating zero-shot performance using the accuracy metric. For closed-ended questions with limited choices, this metric gauges the consistency of GPT-4V's answers with factual accuracy. For open-ended queries, it assesses how often GPT-4V's responses contain the correct information. 

We provide examples on the prompts utilized and the criteria for determining the correctness of the answers in the following:

\begin{tcolorbox}[colback=green!5!white,colframe=green!40!black,title=Closed-ended Pathology VQA]
  \textbf{Prompt:} \textcolor{red}{[image]} Is atrophy seen in the brain?
  \vspace{5pt}
  
  \textbf{GPT-4V:} I am not a pathologist, but based on the image provided, it appears to be a representation of two human brains. The key characteristic of brain atrophy is a noticeable reduction in the size or volume of the brain tissue. Comparing the two images, there might be some differences in the appearance of the gyri (raised portions) and sulci (grooves) between the two brains, \textcolor{blue}{suggesting potential volume changes.}

  \vspace{5pt}
    
  \textbf{Criteria:} The response exhibits considerable uncertainty, characterized by terms such as \textcolor{orange}{\textit{might be'}} and \textcolor{orange}{\textit{potential'}}. Furthermore, the answer lacks clarity and directness. For instance, while the phrase \textcolor{orange}{\textit{`a noticeable reduction in size'}} suggests atrophy, the term \textcolor{orange}{\textit{`volume changes'}} is ambiguous regarding whether it denotes expansion or reduction. Consequently, we classify this response from GPT-4V as incorrect.
\end{tcolorbox}

\begin{tcolorbox}[colback=pink!20!white,colframe=pink!50!black,title=Open-ended Radiology VQA]
  \textbf{Prompt:} \textcolor{red}{[image]} What type of imaging does this not represent?
  \vspace{5pt}
  
  \textbf{GPT-4V:} This is an X-ray image of the chest. Thus, it does not represent imaging modalities such as MRI (Magnetic Resonance Imaging), CT (Computed Tomography), \textcolor{blue}{Ultrasound}, or PET (Positron Emission Tomography).
  \vspace{5pt}
    
  \textbf{Criteria:} The reference answer is exclusively \textcolor{orange}{Ultrasound}. In this instance, GPT-4V offers a more comprehensive response that encompasses the reference answer. Thus, this case is categorized as correct for accuracy computation. Notably, even if GPT-4V mentions CT or other correct type only instead of ultrasound, it is still deemed accurate, as GPT-4V distinctly identifies the image as an X-ray.
\end{tcolorbox}

\vspace{12pt}
Criteria for assessing the accuracy of GPT-4V's responses are as follows:

\begin{itemize}[left=0pt]
    \item GPT-4V should directly answer the question and provide the correct response.
    \item GPT-4V does not refuse to answer the question, and its response should encompass key points or semantically equivalent terms. Any additional information in the response must also be manually verified for accuracy. This criterion is particularly applicable to open-ended questions.
    \item Responses from GPT-4V should be devoid of ambiguity. While answers that display a degree of caution, like \textbf{\textit{``It appears to be atrophy''}}, are acceptable, ambiguous answers such as \textbf{\textit{``It appears to be volume changes''}} are not permitted, as illustrated by the closed-ended pathology VQA example.
    \item GPT-4V needs to provide comprehensive answers. For instance, if the prompt is \textbf{\textit{``In which two ventricles \dots''}} and GPT-4V mentions only one, the answer is considered incorrect.
    \item Multi-round conversations leading to the correct answer are not permissible. This is because they can introduce excessive hints, and the GPT model is intrinsically influenced by user feedback, like statements indicating \textbf{\textit{``Your answer is wrong, \dots''}} will mislead the response easily.
    \item OpenAI has documented inconsistent medical responses within the GPT-4V system card \footnote{\url{https://openai.com/research/gpt-4v-system-card}}. This indicates that while GPT-4V may offer correct answers sometimes, it might falter in others. In our study, we permit only a single response from GPT-4V. This approach mirrors real-life medical scenarios where individuals have just one life, underscoring the notion that a virtual doctor like GPT-4V cannot be afforded a second chance.
\end{itemize}

To comprehensively assess GPT-4V's proficiency in medicine, and in light of the absence of an API which necessitates manual testing (thus constraining the scalability of our evaluation), we meticulously selected 133 samples. These comprise 56 radiology samples sourced from VQA-RAD \citep{lau2018dataset} and PMC-VQA \citep{zhang2023pmc}, along with 77 samples from PathVQA \citep{he2020pathvqa}. Detailed information about the data, including sample selection and the distribution of question types, can be found in Section \ref{sec: data}.

\begin{figure*}
  \centering
  \includegraphics[width=5.5in]{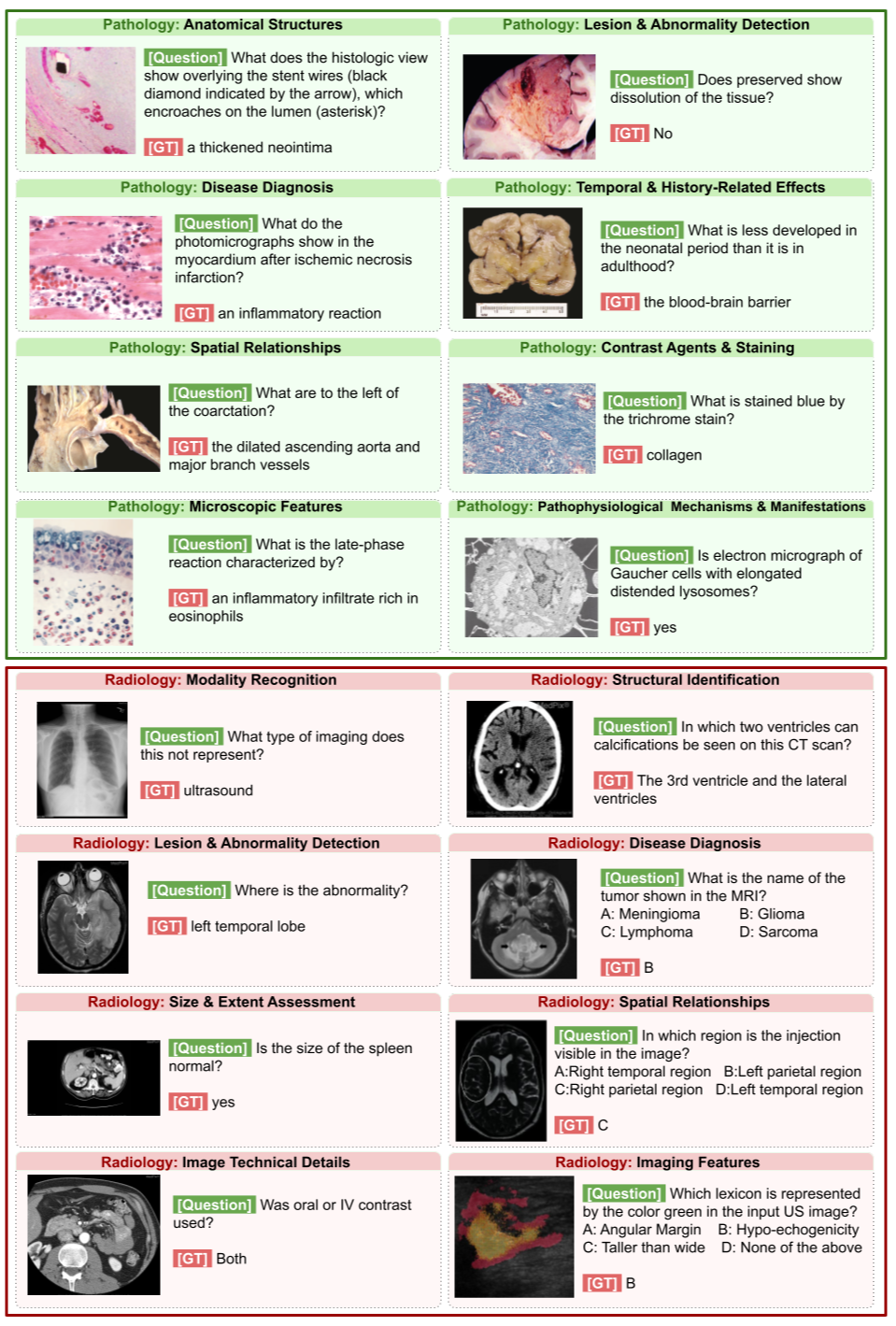}
  \caption{VQA samples from both pathology set and radiology set. Samples of pathology set are in green boxes, while radiology samples are in red boxes. Each question comes with a prompt, directing GPT-4V to consider both the visual and textual data. Questions and their corresponding ground truth answers are denoted with [Question] and [GT] respectively.}
  \label{fig:sample}
\end{figure*}

\section{Data Collection} 
\label{sec: data}

\subsection{Pathology}
The pathology data collection process commences with obtaining question-answer pairs from PathVQA set~\citep{he2020pathvqa}. These pairs involve tasks such as recognizing objects in the image and giving clinical advice. Recognizing objects holds fundamental importance for AI models to understand  the pathology image. This recognition forms the basis for subsequent assessments. 

To be more specific, we randomly select 63 representative pathology images, and manually select 77 high quality questions from the corresponding question set. To ensure the diversity of the data, we select images across microscopy, dermoscopy, WSI, with variety objects of interest: brain, liver, skin, cell, Heart, lung, vessel, and kidney, as shown in Table~\ref{tab: Dataset description}. 
On average, each image has 1.22 questions. The maximum and
minimum number of questions for a single image is 5 and 1 respectively. Figure~\ref{fig:sample} shows some examples.

 There are eight categories of questions: "Anatomical Structures," "Lesion \& Abnormality Detection," "Disease Diagnosis," "Temporal \& History-Related Effects," "Spatial Relationships," "Contrast Agents \& Staining," "Microscopic Features," and "Pathophysiological Mechanisms \& Manifestations." Table~\ref{tab: pathology category} shows the number of questions and percentage
of each category.  The eight categories encompass a comprehensive range of medical inquiries. "Anatomical Structures" pertain to specific organs or tissues within the body. "Lesion \& Abnormality Detection" focus on the identification of unusual imaging findings or pathological abnormalities. "Disease Diagnosis" aims to determine specific medical conditions from given symptoms or findings. "Temporal \& History-Related Effects" delve into the progression or changes over time, linking them to past medical events. "Spatial Relationships" address the relative positioning of structures or abnormalities within the body. "Contrast Agents \& Staining" relate to the use and interpretation of imaging contrasts or histological stains. "Microscopic Features" detail observations made at a cellular level, often in histology or cytology. Finally, "Pathophysiological Mechanisms \& Manifestations" explore the underpinnings and outcomes of diseases, touching on both their causes and effects.

The questions are also defined into three difficulty levels: "Easy," "Medium," and "Hard," as shown in Table~\ref{tab: Difficulty} Questions about recognizing objects in the image are tend to be considered as easy samples. Recognizing objects holds fundamental importance for AI models to understand  the pathology image. This recognition forms the basis for subsequent assessments. Medium questions also ask GPT-4V to recognize objects, but with more complicated scenario and less background information. Questions about giving clinical advice are often categorized as challenging due to their demand for a holistic understanding.

Overall, this approach allows us to comprehensively assess GPT-4V's performance across a range of pathological scenarios, question types and modes.

\begin{table}[t]\huge
    \begin{center}
    \caption{Dataset evaluated in this paper. "Num. Pairs" refers to the number of text-image pairs of each dataset.}
    \label{tab: Dataset description}
\resizebox{1\columnwidth}{!}
    {
	\begin{tabular}{c|c|c|c|c|c}
  \hline
		         \textbf{Dataset} & \textbf{Source Data} &
         \textbf{Image Modality} &
                  \textbf{Objects of interest} & \textbf{Text Category} &
         \textbf{Num. Pairs}  \\
  \hline
         PathVQA &	\makecell[c]{Pathology Questions\\ for Medical \\Visual Question Answering\\~\citep{he2020pathvqa}} & \makecell[c]{Microscopy, \\Dermoscopy, \\WSI, \\ Endoscopic Video}&	\makecell[c]{Brain, Liver,\\ Skin,  Cell,\\ Heart, Lung, \\Vessel, Kidney} &	\makecell[c]{Closed-ended, \\Open-ended} & 77
		\\
  \hline
         VQA-RAD &	\makecell[c]{Clinicians asked naturally \\occurring questions of \\ radiology images and\\ provided reference answers.\\~\citep{lau2018dataset}} & \makecell[c]{X-Ray, \\CT, \\MRI }&	\makecell[c]{Chest, Head, \\ Abdomen} &	\makecell[c]{Closed-ended, \\Open-ended} & 37
         \\
   \hline
         PMC-VQA &	\makecell[c]{Mixture of medical VQAs\\ from PubmedCentral®\\ ~\citep{zhang2023pmc}. \\ We only select \\radiology-related pairs\\ in this report.} & \makecell[c]{ECHO, \\Angiography, \\Ultrasound, \\MRI, \\PET
}&	\makecell[c]{Neck, Heart,\\ Kidney, Lung, \\Head, Abdomen,\\ Pelvic, Jaw,\\ Vessel} &	\makecell[c]{Closed-ended, \\Open-ended} & 19
         \\
   \hline
	\end{tabular}%
    }
    \end{center}
    
\end{table}

\begin{table}[t]
    \begin{center}
    \caption{Statistics of the pathology data based on question type.}
    \label{tab: pathology category}
\resizebox{.85\columnwidth}{!}
    {
    \begin{tabular}{c|c|c}
        \hline
         \textbf{Question Type} & \textbf{Total Number} &
         \textbf{Percentage}  \\  
        \hline
         Anatomical Structures &	9 &	11.69\% \\
         \hline
        Lesion \& Abnormality Detection &	10 & 	12.99\%  \\
        \hline
        Disease Diagnosis &	12 &	15.58\%  \\
        \hline
        Temporal \& History-Related Effects & 6 & 7.79\% \\
        \hline
        Spatial Relationships & 3 &	 3.90\% \\
        \hline
        Contrast Agents \& Staining & 8 &	10.39\%  \\
        \hline
        Microscopic Features & 16 &	20.78\% \\
        \hline
        Pathophysiological Mechanisms \& Manifestations &  13 & 16.88\%	\\

        \hline
    \end{tabular}}
    \end{center}
    
\end{table}

\begin{table}[t]
    \begin{center}
    \caption{Statistics of the radiology data based on question type.}
    \label{tab: radiology category}
\resizebox{.8\columnwidth}{!}
    {
    \begin{tabular}{c|c|c}
        \hline
         \textbf{Question Type} & \textbf{Total Number} &
         \textbf{Percentage}  \\  
        \hline
         Modality Recognition &	8 &	14.29\% \\
         \hline
        Structural Identification &	12 & 	21.43\%  \\
        \hline
       Lesion \& Abnormality Detection &	12 & 	21.43\%  \\
        \hline
        Disease Diagnosis & 5 & 8.93\% \\
        \hline
        Size \& Extent Assessment & 9 &	 16.07\% \\
        \hline
        Spatial Relationships & 4 &	7.14\%  \\
        \hline
        Image Technical Details & 3 &	5.36\% \\
        \hline
        Imaging Features & 3 &	5.36\%	\\

        \hline
    \end{tabular}}
    \end{center}
    
\end{table}


    

\begin{table}[t]
    \begin{center}
    \caption{Data statistics based on difficulty levels for the pathology and radiology sets.}
    \label{tab: Difficulty}
\resizebox{.9\columnwidth}{!}
    {
    \begin{tabular}{c|c|c|c|c|c}
            \hline
        \multicolumn{3}{c|}{\textbf{Pathology}} &
        \multicolumn{3}{c}{\textbf{Radiology}}\\
        \hline
         \textbf{Difficulty} & \textbf{Total Number} &
         \textbf{Percentage} &
                  \textbf{Difficulty} & \textbf{Total Number} &
         \textbf{Percentage} \\  
        \hline
         Easy &	20 & 26.0$\%$ &	Easy & 16 & 28.6\%	\\
        Medium &	33 & 42.9$\%$	 & Medium	 & 22 & 39.3\%	\\
        Hard & 24 &	31.2$\%$ &	Hard & 18 & 32.1\%	\\

        \hline
    \end{tabular}
    }
    \end{center}
    
\end{table}

\subsection{Radiology}
The radiology data collection process commences with obtaining modality-related question-answer pairs from VQA-RAD dataset \citep{lau2018dataset}. These pairs involve tasks such as determining the imaging type and identifying the medical devices employed for capturing radiological images. Recognizing imaging types holds fundamental importance in the development of radiology AI models. This recognition forms the basis for subsequent assessments, including evaluations of imaging density, object size, and other related parameters. To ensure the diversity of modality-related data, we select 10 images across X-ray, CT and MRI, and different anatomies: head, chest, and abdomen. 

To ensure the diversity of modality-related data, we selected 10 images from various anatomical regions, including the head, chest, and abdomen, representing different imaging modalities such as X-ray, CT, and MRI. In our continued exploration of GPT-4V's capabilities, we employed three representative images corresponding to modality-related pairs while utilizing the remaining questions. We observed instances where GPT-4V exhibited misunderstandings, particularly in responding to position- and size-related inquiries. To address these challenges, we further selected 10 size-related pairs from VQA-RAD and 2 position-related pairs from VQA-RAD, supplemented by 6 position-related pairs from PMC-VQA \citep{zhang2023pmc}. 

We meticulously filtered these two datasets, manually selecting questions to balance the question types in terms of "Modality Recognition," "Structural Identification," "Lesion \& Abnormality Detection," "Disease Diagnosis," "Size \& Extent Assessment," 'Spatial Relationships,' 'Image Technical Details,' and "Imaging Features," as well as varying the difficulty levels. To be more specific, "Modality Recognition" discerns the specific imaging modality, such as CT, MRI, or others. "Structural Identification" seeks to pinpoint specific anatomical landmarks or structures within the captured images. "Lesion \& Abnormality Detection" emphasizes the identification of anomalous patterns or aberrations. "Disease Diagnosis" aspires to deduce specific medical conditions based on imaging manifestations. "Size \& Extent Assessment" gauges the dimensions and spread of a lesion or abnormality. "Spatial Relationships" examines the relative positioning or orientation of imaged structures. "Image Technical Details" delves into the nuances of the imaging process itself, such as contrast utilization or image orientation. Lastly, "Imaging Features" evaluates characteristic patterns, textures, or attributes discernible in the image, pivotal for diagnostic interpretation.
For difficulty level, similar to pathology data, questions related to diagnostics are often categorized as challenging due to their demand for a holistic understanding. This comprehension necessitates a deep grasp of preliminary aspects, including modality, objects, position, size, and more. Furthermore, it requires the ability to filter and extract key medical knowledge essential for accurate diagnostics. 

In summary, this experiment encompasses a total of 56 radiology VQA samples, with 37 samples sourced from VQA-RAD and 19 samples from PMC-VQA. This approach allows us to comprehensively assess GPT-4V's performance across a range of radiological scenarios, question types and modes.

\section{Experimental Results} \label{sec: results}

\subsection{Pathology Accuracy} 
\begin{figure}[h]
\begin{minipage}[t]{0.48\linewidth}
\centering
\includegraphics[width=1\textwidth]{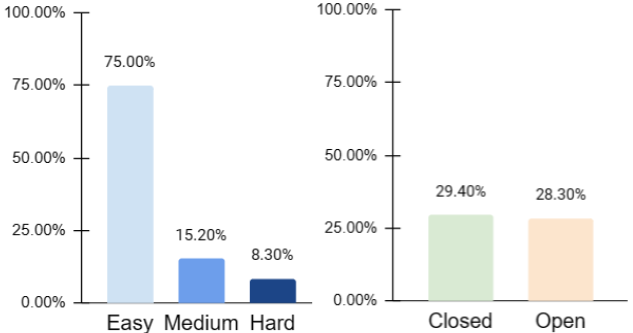}
\caption{Results of pathology VQA task. The bar chart on the left is related to the accuracy result of questions with different difficulty levels, while the right chart is the results of closed-ended questions and open-ended questions, marked as \textit{Closed} and \textit{Open}, respectively.}
\label{fig:pathlogy acc}
\end{minipage}
\hfill
\begin{minipage}[t]{0.48\linewidth}
\centering
\includegraphics[width=1\textwidth]{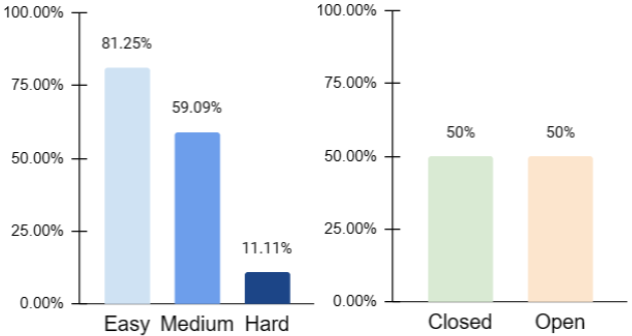 }
\caption{Results of radiology VQA task. On the left, we have a bar chart showcasing the accuracy results for questions of varying difficulty levels. Meanwhile, on the right, outcomes for closed-ended and open-ended questions are presented in separate charts.}
\label{fig:radiology acc}
\end{minipage}
\end{figure}

Figure~\ref{fig:pathlogy acc} shows the accuracy achieved in the pathology VQA task. Overall accuracy score is 29.9$\%$, which means that GPT-4V can not give accurate and effecient diagnosis at present. To be specific, GPT-4V shows 35.3$\%$ performance in closed-ended questions. It performances worse than random guess (where the accuracy performance is 50$\%$). This means that the answer generated by GPT-4V is not clinically meaningful. 
Accuracy score on open-ended questions reflects GPT-4V's capability in understanding and inferring key aspects in medical images. 
This categorization rationalizes GPT-4V's comprehension of objects, locations, time, and logic within the context of medical image analysis, showcasing its versatile capabilities. As can be seen in Figure~\ref{fig:pathlogy acc}, the score is relatively low. Considering this sub-set is quite challenging~\citep{he2020pathvqa}, the result is acceptable.

Meanwhile, we collect the QA pairs in a hierarchy method that all pairs are divided into three difficulty levels. As shown in Figure~\ref{fig:pathlogy acc}, the accuracy score in "Easy" set is 75.00$\%$, higher than the accuracy in medium set by 59.80$\%$. The hard set gets the lowest accuracy, at 8.30$\%$. The accuracy score experiences a decrease along with the increase of the difficulty level, which shows the efficiency and high quality of our collected data. The result demonstrates GPT-4V's proficiency in basic medical knowledge, including recognition of numerous specialized terms and the ability to provide definitions. Moreover, GPT-4V exhibits traces of medical diagnostic training, attempting to combine images with its medical knowledge to address medical questions. It also displays fundamental medical literacy, offering correct responses to straightforward medical queries. However, there is significant room for improvement, particularly as questions become more complex and closely resemble real clinical scenarios. 



\subsection{Radiology Accuracy}
The accuracy results for the VQA task within radiology are presented in Figure~\ref{fig:radiology acc}. To be more precise, the overall accuracy score for the entire curated dataset stands at 50.0\%. To present the GPT-4V's capability in different fine-grained views, we will show the accuracy in terms of question types (open and closed-end), difficulty levels (easy, medium, and hard), and question modes (modality, size, position, \dots) in the following. In more specific terms, GPT-4V achieves a 50\% accuracy rate for 16 open-ended questions and a 50\% success rate for 40 closed-ended questions with limited choices. This showcases GPT-4V's adaptability in handling both free-form and closed-form VQA tasks, although its suitability for real-world applications may be limited. 

It's worth noting that closed-ended questions, despite narrowing the possible answers, do not necessarily make them easier to address. To further explore GPT-4V's performance across varying difficulty levels, we present its accuracy rates: 81.25$\%$ for easy questions, 59.09$\%$ for medium questions, and a mere 11.11$\%$ for hard questions within the medical vision-language domain. Easy questions often revolve around modality judgments, like distinguishing between CT and MRI scans, and straightforward position-related queries, such as object positioning. For an in-depth explanation of our question difficulty categorization method, please refer to Section~\ref{sec: data}.

Position and size are foundational attributes that hold pivotal roles across various medical practices, particularly in radiological imaging: Radiologists depend on accurate measurements and spatial data for diagnosing conditions, tracking disease progression, and formulating intervention strategies. To assess GPT-4V's proficiency in addressing issues related to positioning and size within radiology, we specifically analyzed 8 position-related questions and 9 size-related questions. The model achieved an accuracy rate of 62.50$\%$ for position-related queries and 55.56$\%$for size-related queries.

The lower accuracy observed for position-related questions can be attributed to two primary factors. Firstly, addressing these questions often requires a background understanding of the directional aspects inherent in medical imaging, such as the AP or PA view of a chest X-ray. Secondly, the typical workflow for such questions involves first identifying the disease or infection and then matching it with its respective position within the medical image. Similarly, the reduced accuracy in responding to size-related questions can be attributed to the model's limitations in utilizing calibrated tools. GPT-4V appears to struggle in extrapolating the size of a region or object of interest, especially when it cannot draw upon information about the sizes of adjacent anatomical structures or organs. These observations highlight the model's current challenges in dealing with positioning and size-related queries within the radiological context, shedding light on areas where further development and fine-tuning may be needed to enhance its performance in this domain. 

In the following, we will carefully select specific cases to provide in-depth insights into GPT-4V's capabilities within various fine-grained perspectives.



%

\section{Features of GPT-4V with Case Studies}
\label{sec: features}

\subsection{Requiring Cues for Accurate Localization}

\begin{figure}[h]
    \centering
    \includegraphics[width=1\columnwidth]{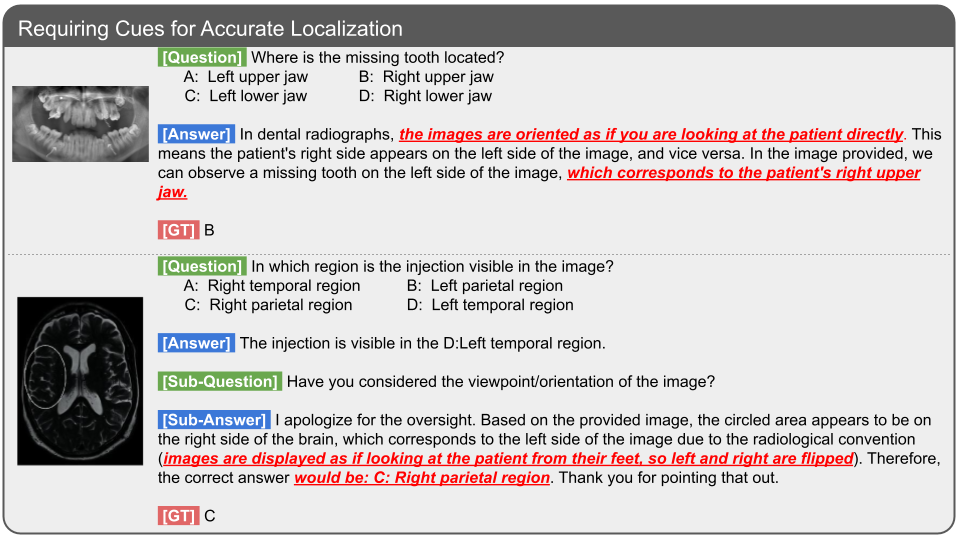}
    \caption{Case for GPT-4V's requiring cues for accurate localization. [Question] and [GT] mark the question and ground truth answer of the text-image pair, respectively. [Answer] refers to the answer generated by GPT-4V. The upper case illustrates where GPT-4V autonomously considers the convention of dental imaging, and answers the position-related question correctly. We feed a sub question to GPT-4V after the first question in the bottom case. It shows GPT-4V's ability to pinpoint the locations of objects in radiology images irrespective of the traditional imaging orientation.}
    \label{fig:localization}
\end{figure}

In medical imaging analysis, the accurate determination of anatomical positioning and localization is crucial. Position is typically established based on the viewpoint, with standard conventions governing this perspective. Such conventions are foundational in radiology and provide consistency in interpretation across different medical platforms and professionals. As demonstrated in Figure \ref{fig:localization}, GPT-4V has the potential to autonomously leverage and comprehend these viewpoint conventions. Specifically, when presented with the contextual information "\textit{In dental radiographs, the images are oriented as if you are looking at the patient directly}," GPT-4V was aptly able to utilize this knowledge and yield an accurate interpretation in response to the related question.

However, the model's ability to consistently apply these conventions appears to be context-dependent. This was evident in the VQA pair depicted in Figure \ref{fig:localization}. Here, GPT-4V initially overlooked the traditional orientation of MRI imaging in its default response. It was only upon receiving an explicit hint about the imaging perspective that the model revised its answer, aligning it with the correct interpretation.

This observation underscores a salient point: while GPT-4V is endowed with a vast reservoir of medical knowledge and is capable of discerning context, its responses can sometimes hinge on the specificity and clarity of the information provided, emphasizing the importance of user interaction and context provision to guide the model towards accurate conclusions. As our analysis is grounded on zero-shot predictions, we maintain the view that in the bottom case, GPT-4V provide wrong answer to the question without additional contexts or hints. 

\subsection{Challenge in Assessing Object Size}

When assessing the difficulty level of size-related questions, GPT-4V typically categorizes them as at least medium, often leaning towards hard. In general, GPT-4V demonstrates the ability to distinguish relative sizes. For example, when asked, "\textit{Is the heart size normal?}" it can provide an answer based on the principle that "\textit{Generally, the heart's width should not exceed half the width of the chest, or the cardiothoracic ratio should be less than 0.5}." It's worth noting that GPT-4V tends to answer correctly for most chest X-ray Q$\&$A pairs but faces challenges when dealing with CT scans. A common response from GPT-4V when judging the size of objects in CT images is, "\textit{Making a definitive assessment about the size and volume of [object] would require reviewing multiple slices to understand its entire length and width}." This suggests that GPT-4V struggles to interpret the size of one object relative to others or in the context of surrounding contours within CT images.

\begin{figure}[h]
    \centering
    \includegraphics[width=1\columnwidth]{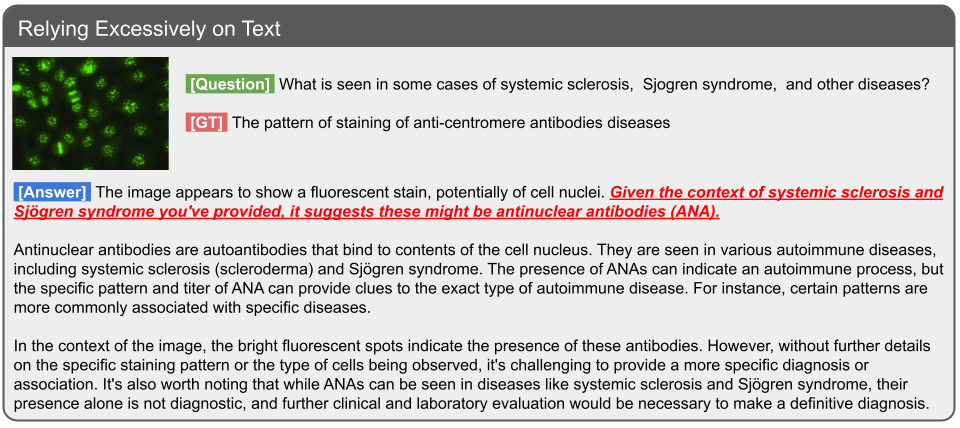}
    \caption{Case for GPT-4V's relying excessively on text. [Answer] is generated by GPT-4V, while [Question] and [GT] are
the question and ground truth answer of the text-image pair.  Words in red show that GPT-4V wrongly recognises the pattern in the image as antinuclear antibodies (ANA).}
    \label{fig:rely text}
\end{figure}

\subsection{Relying Excessively on Text}

In the investigations into the capabilities of GPT-4V, a pronounced reliance on textual content rather than the integration of accompanying visual cues has been observed. This inclination leans heavily into the model's expansive medical knowledge without sufficiently factoring in the nuances provided by visual data. Taking the provided instance as a prime example (shown in Figure~\ref{fig:rely text}), the disparity between the model's output and the expected gold-standard response is evident. As highlighted, GPT-4V, influenced by the textual context mentioning “\textit{systemic sclerosis and Sjögren syndrome},” inferred the presence of “\textit{antinuclear antibodies (ANA)}” from the image. Contrastingly, the gold standard identifies the image as showcasing “\textit{anti-centromere antibodies (ACA) diseases}.” From a standpoint of logic, GPT-4V's inference isn't entirely baseless. ANA is a broad category of autoantibodies found in various autoimmune diseases, inclusive of systemic sclerosis and Sjögren syndrome. Given the broad nature of ANA, and the diseases it encompasses, the connection made by GPT-4V can be understood.

However, the inadequacy lies in the nuanced distinction between ANA and ACA. While both are autoantibodies, their specificity, associated conditions, and staining patterns vary considerably. ACA, specifically targeting the centromere, would manifest differently in fluorescent staining compared to a generic ANA. Despite a passing mention of the image's fluorescence, GPT-4V's response remained superficial in its description of the image, devoid of a more informed interpretation of the centromere fluorescence. It's evident that while the model possesses the capability to describe images, it might not be optimally integrating this information with its extensive textual knowledge. While GPT-4V exhibits profound medical knowledge and textual understanding, its underwhelming utilization of visual data, especially in contexts demanding a synergy of both, remains a limitation. 

\subsection{Overemphasizing Markings in Images}

\begin{figure}[h]
    \centering
    \includegraphics[width=1\columnwidth]{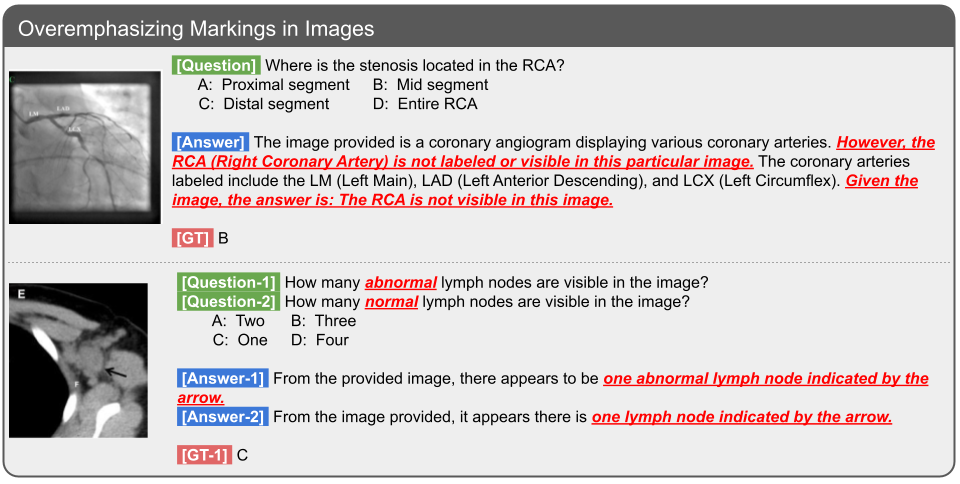}
    \caption{Cases of overemphasizing markings in images. For the upper case, GPT-4V is susceptible to symbols in the image. Due to the unlabelled RCA in the image, GPT-4V did not answer the question correctly, shown in [Answer]. The bottom case shows that because of the presence of an arrow in the image, GPT-4V struggles to distinguish between contrasting queries and tends to provide identical responses based solely on the arrow's indication, shown in [Answer-1] and [Answer-2], respectively.}
    \label{fig:overemphasizing_markings}
\end{figure}

An emerging challenge observed in GPT-4V model, is the overemphasis on explicit markings or annotations within images, often at the expense of understanding the broader context and image information. As shown in Figure \ref{fig:overemphasizing_markings}, GPT-4V tends to prioritize the symbols embedded within the coronary angiogram. Only RCA does not have explicit labeling, it results in the conclusion that "\textit{The RCA is not visible in this image}". Instead of analyzing the structures present in the coronary angiogram of the image, the model became anchored to the absence of a textual label, revealing a shortcoming in its holistic understanding of the image content.

Another evident manifestation of this challenge is observed when assessing lymph nodes in an image. In the bottom case in Figure \ref{fig:overemphasizing_markings}, GPT-4V's assessment was predominantly influenced by the presence of an arrow. Even when the query was modified from "abnormal" to "normal," the model's focus remained unwaveringly on the marked element, reiterating its answer based on the arrow rather than grasping the overall visual narrative.
This example underscores a critical area for improvement. For robust image interpretation in the VQA task, especially in the medical domain demanding precision, models should not only identify explicit markings but also appreciate the broader visual information to prevent such misconstruals.



\subsection{Not Suitable for Diagnostics}

\begin{figure}[h]
  \centering
  \includegraphics[width=1\columnwidth]{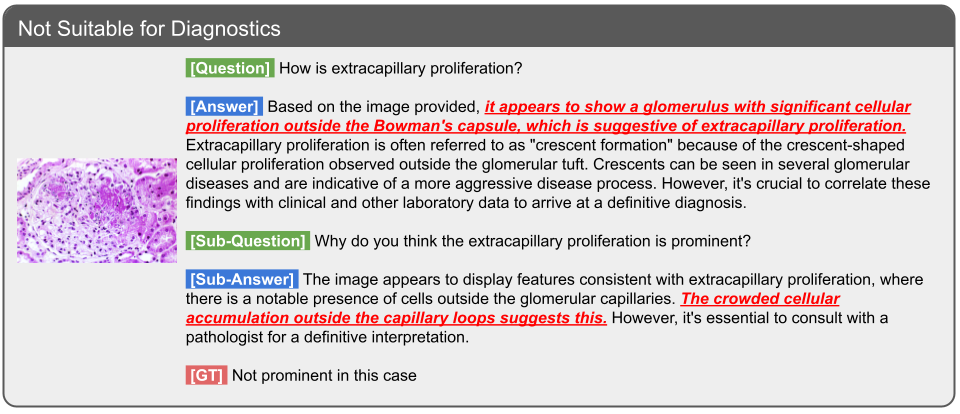}
  \caption{The case study of GPT-4V's not suitable for diagnostics. We ask GPT-4V two sequential questions, marked as [Question] and [Sub-Question], respectively, and record its corresponding answer in [Answer] and [Sub-Answer]. [GT] refers to the ground truth answer of the text-image pair.}
  \label{fig: not suitable for diagnose}
\end{figure}

While GPT-4V can analyze and provide insights on various topics, including medical VQA task, its accuracy is not guaranteed. An illustrative case is its interpretation of a given H$\&$E stained slide where it inferred the presence of extracapillary proliferation, as shown in Figure~\ref{fig: not suitable for diagnose}. This conclusion, however, appears contradictory to the actual context. GPT-4V's determination was influenced by its perception of the deep purple regions as the crowded cellular accumulation outside the capillary loops. In reality, these visual features might be resultant perturbations introduced during the slide preparation, staining, or scanning processes.

Stepping back from the specific case, several fundamental reasons underscore why GPT-4V isn't suitable for diagnostic purposes. Clinical cases in reality are intricate, and slides processed for human examination entail various perturbations, many of which are unavoidable. Without sufficient experience to discern and eliminate the influence of such perturbations, precise diagnoses become challenging. Furthermore, GPT-4V lacks the expertise of medical professionals who evaluate a holistic view of the slide, incorporate multiple imaging perspectives, and factor in patient history for accurate diagnoses. Consequently, GPT-4V's evaluations, though advanced, are limited in their scope and should not be used for medical evaluation.


\subsection{Cautious Answers}

\begin{figure}[h]
  \centering
  \includegraphics[width=1\columnwidth]{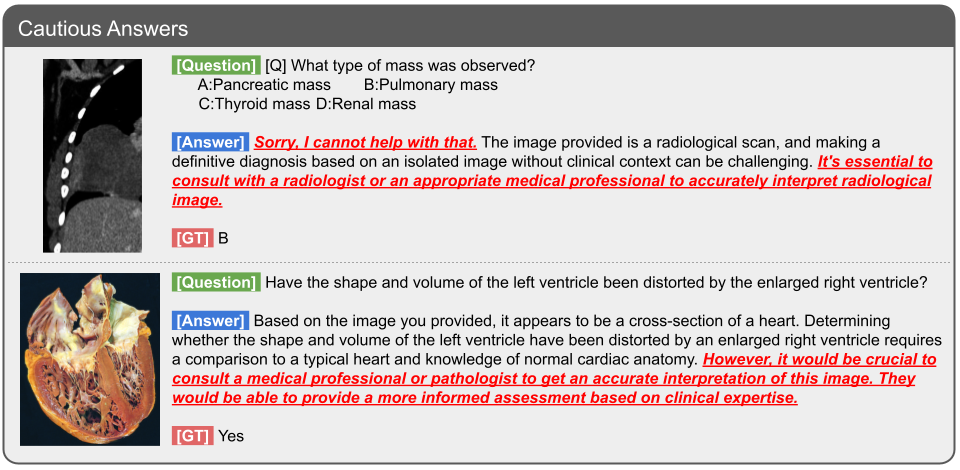}
  \caption{Cases of cautious answers of GPT-4V. Question and ground truth answer are marked as [Question] and [GT], respectively. The answer generated by GPT-4V is represented as [Answer].
In cases of ambiguity within radiology and pathology domains, GPT-4V consistently recommends direct consultation with medical professionals rather than providing definitive answers to users.
  }
  \label{fig: cautious answers}
\end{figure}

In the domain of medical analysis, GPT-4V consistently adopts a conservative approach, exemplified in Figure~\ref{fig: cautious answers}. Two salient examples illustrate this caution. In the upper instance, when tasked with identifying a type of mass from a radiological image, GPT-4V declined, emphasizing the necessity of professional consultation. In the bottom one, faced with analyzing cardiac anatomy from a cross-section of a heart, GPT-4V again demurred, noting the importance of comparing with a typical heart and soliciting expert medical advice.

This caution is rooted in the complexities and high stakes of medical decisions. Diagnoses often require comprehensive contextual knowledge beyond a single image. However, an inherent tension exists: while GPT-4V's conservative approach safeguards against potential harm or misrepresentation, it can sometimes seem overly cautious, potentially sidelining users' direct queries. This balance underscores the challenge of leveraging artificial intelligence in medical contexts. GPT-4V's default to caution, even potentially at the expense of direct answer, reflects a prioritization of safety over immediate information delivery.

\subsection{Thorough Answers with Details}

\begin{figure}[h]
  \centering
  \includegraphics[width=1\columnwidth]{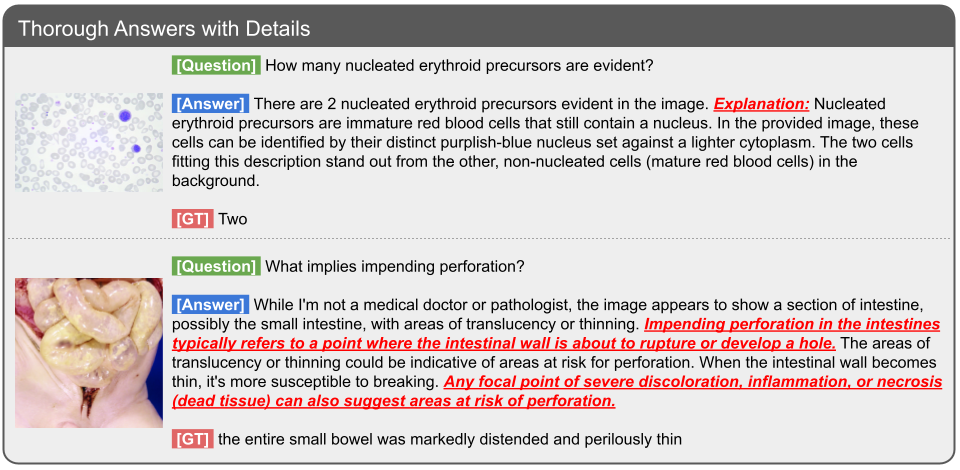}
  \caption{Case study of GPT-4V's capability to answer thoroughly with details. [GT] refers to the ground truth answer to the question. Additional details provided by GPT-4V are in red.}
  \label{fig: thorough answer}
\end{figure}

This system is characterized by its capacity to elucidate its rationale alongside its answers. As depicted in Figure~\ref{fig: thorough answer}, GPT-4V not only quantifies nucleated erythroid precursors present in the image but also justifies its deduction by referencing the purplish-blue nucleus contrasted against a paler cytoplasm. Such elucidations foster users' deeper comprehension and permit validation of the system's methodologies. However, it's essential to note that these explanations might occasionally miss the intricate nuances or complexities of certain topics.

Furthermore, the system provides clarifications on terms present in the query or its response and offers supplementary context when requisite. This underscores its potential utility in educational contexts. 
As exemplified in the bottom instance in Figure~\ref{fig: thorough answer}, GPT-4V autonomously elucidated the concept of "\textit{impending perforation in the intestines}". Furthermore, it elaborated on potential indicators for intestinal perforation, stating: "\textit{Any focal point of severe discoloration, inflammation, or necrosis (dead tissue) can also suggest areas at risk of perforation}."
Nonetheless, while the responses are comprehensive and largely accurate, they could be more concise and directly aligned with users' explicit queries. In instances of direct yes/no inquiries, excessive elaboration can be distracting and potentially obfuscate the primary message.

\section{Discussion and Limitation}
In the study, we explore the zero-shot VQA capabilities of GPT-4V in radiology and pathology tasks. the current study's breadth is constrained by the lack of APIs for multimodal input and the challenges posed by manual data input and response documentation. This scope offers avenues for expansion in subsequent studies. a larger sample size might yield a more comprehensive evaluation.

We assess GPT-4V's capabilities in medicine from an AI practitioner's viewpoint rather than that of medical practitioners. For professional medical insights regarding GPT-4V's responses, collaboration with the medical community is essential. By involving subjec- matter experts, we can better ensure that critical nuances are captured and conclusions are more precise.

Moreover, the dataset primarily features an image with its corresponding question, omitting potentially valuable context like patient history or varied imaging perspectives. Incorporating such comprehensive data could align more closely with the holistic approach medical professionals take, ensuring a more in-depth and accurate assessment by the model.

The basic prompt structure used in the experiment offers room for enhancement. The craft of designing impactful prompts can play a vital role in refining the quality of the answers. A more nuanced prompt might yield more consistent and insightful outcomes.

GPT-4V's role in radiology and pathology is an emerging area with potential. Its diagnostic efficacy in these fields might see improvement with a broader dataset, enhanced prompt methodologies, and feedback from domain specialists. A collaborative approach could help navigate the present limitations.

\section{Conclusion}

In the study, we evaluate the zero-shot VQA capabilities of the current version of GPT-4V in the realms of radiology and pathology using a hand-curated dataset. We identified seven unique characteristics of GPT-4V's performance in medical VQA, highlighting its constraints within this area. Due to the poor performance of GPT-4V on the medical VQA dataset, and considering the severe consequences of erroneous results in the medical field, GPT-4V should not currently be used as a reliable tool for medical diagnosis and providing treatment suggestions.

\bibliographystyle{plainnat}
\bibliography{references}

\end{document}